\documentclass[runningheads]{llncs}
\usepackage[T1]{fontenc}
\usepackage{amsmath}
\usepackage{graphicx}
\usepackage{hyperref}
\usepackage{makecell}
\begin{document}
\title{SAda-Net: A Self-Supervised Adaptive Stereo Estimation CNN For Remote Sensing Image Data}
\titlerunning{SAda-Net}
%
\author{Dominik Hirner\inst{1}\and
Friedrich Fraundorfer\inst{1}\inst{2}}
\authorrunning{D. Hirner}
%
\institute{Graz University of Technology, Austria \\ Institute of Computer Graphics and Vision 
\email{office@icg.tugraz.at}\\
\url{https://www.tugraz.at/institute/icg/home}
\email{\{dominik.hirner,fraundorfer\}@icg.tugraz.at} \and Remote Sensing Technology Institute (IMF) \\ German Aerospace Center (DLR)
\url{https://www.dlr.de/en}
\email{contact-dlr@dlr.de}
}
\maketitle              
\begin{abstract}
Stereo estimation has made many advancements in recent years with the introduction of deep-learning. However the traditional supervised approach to deep-learning requires the creation of accurate and plentiful ground-truth data, which is expensive to create and not available in many situations. This is especially true for remote sensing applications, where there is an excess of available data without proper ground truth. 
To tackle this problem, we propose a self-supervised CNN with self-improving adaptive abilities. In the first iteration, the created disparity map is inaccurate and noisy. Leveraging the left-right consistency check, we get a sparse but more accurate disparity map which is used as an initial pseudo ground-truth. This pseudo ground-truth is then adapted and updated after every epoch in the training step of the network.
We use the sum of inconsistent points in order to track the network convergence.
The code for our method is publicly available at: 
\href{https://github.com/thedodo/SAda-Net}{https://github.com/thedodo/SAda-Net} 
\end{abstract}

\section{Introduction}

Stereo Vision has been a major topic of computer vision for many years. The goal of stereo vision is to extract 3D information of a scene using two neighboring images showing the same scene taken from different camera poses. 3D scene information is useful for many important applications, such as robotics, autonomous driving, 3D scene reconstructions or virtual and augmented reality. 3D reconstruction of urban aerial images is particular important for a multitude of use-cases, such as urban planning, environmental monitoring or disaster management and prevention. 

\begin{figure}[t]
  \centering
   \includegraphics[width=0.3\linewidth]{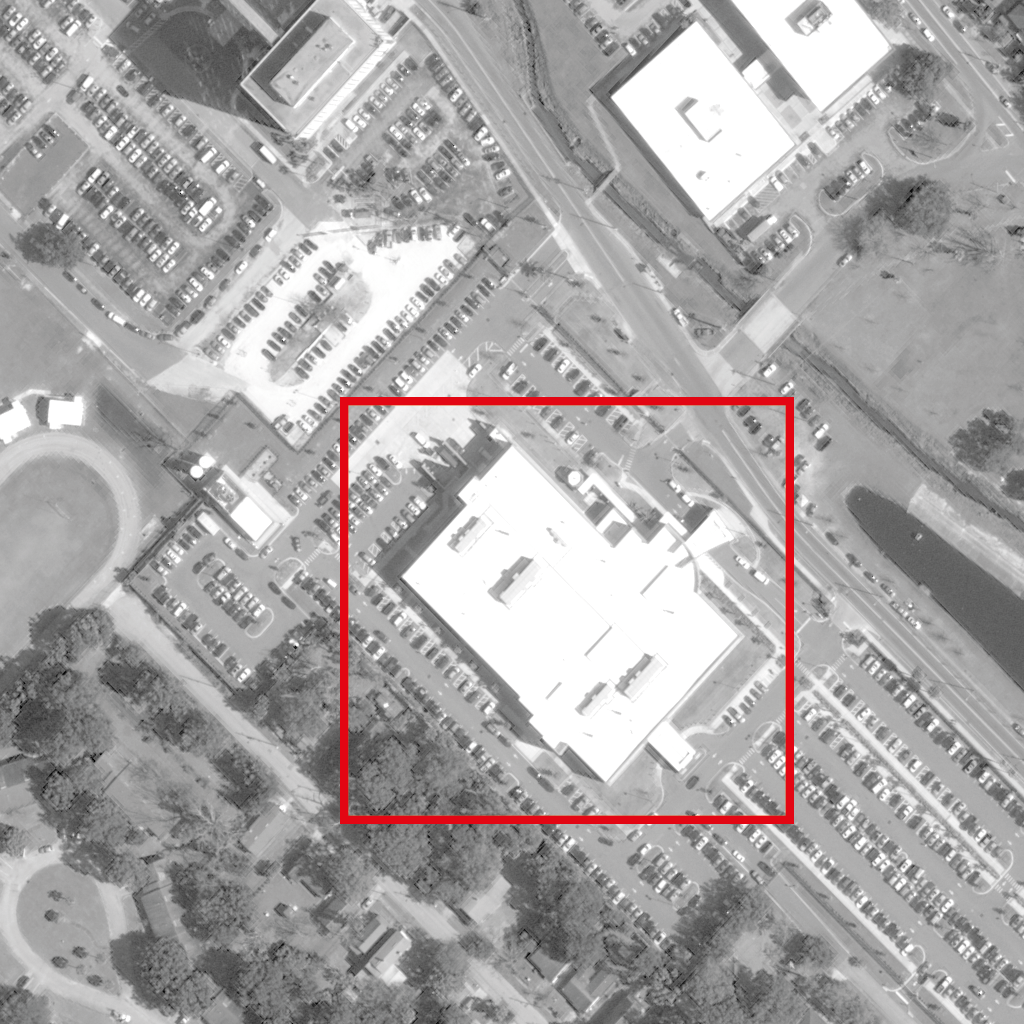}
   \includegraphics[width=0.3\linewidth]{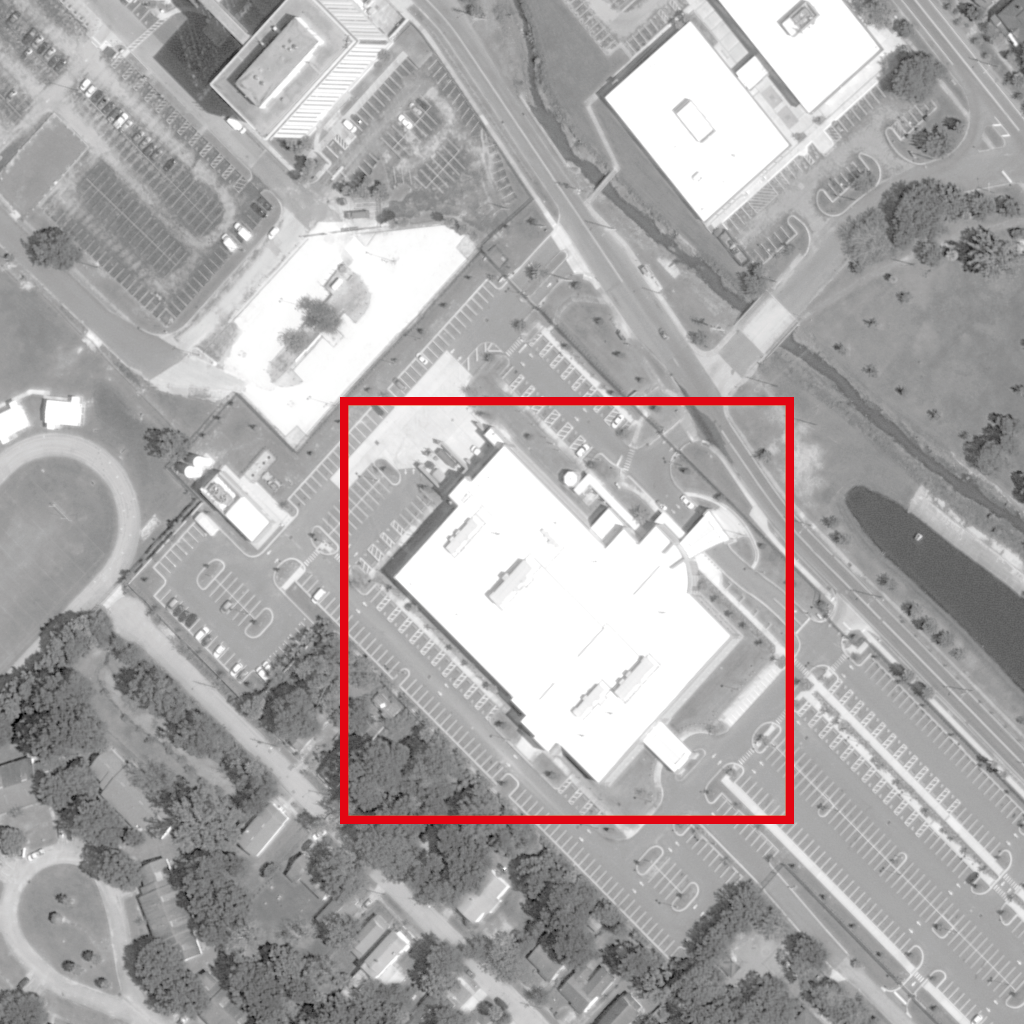}
   \includegraphics[width=0.3\linewidth]{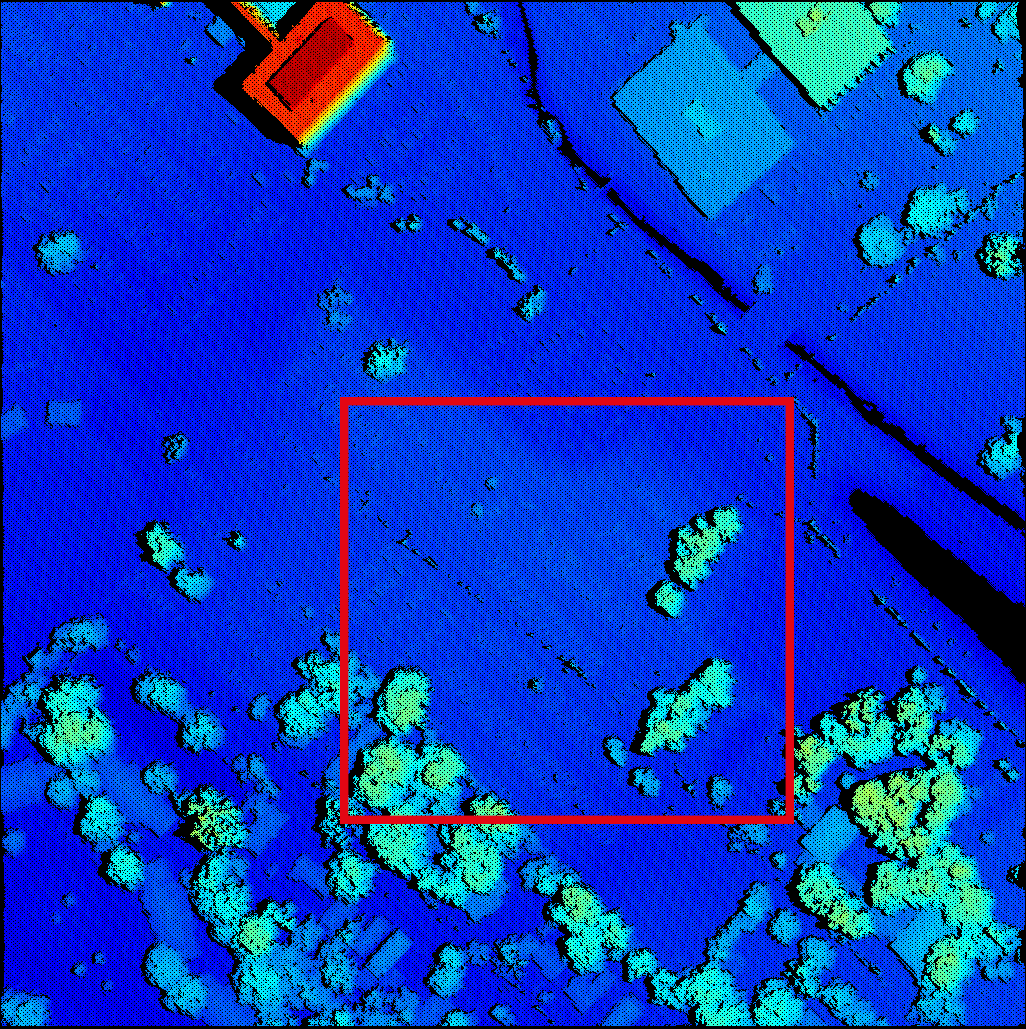}
   \caption{This image shows one example of the 2019 IEEE Data Fusion Contest (DFC2019)~\cite{data_fusion} data set where the scene has changed between capturing the ground truth data and the image data. F.l.t.r.: rectified reference image, rectified second image and ground truth disparity map. The large building visible in the panchromatic images is missing in the disparity map.}
    \label{fig:missing_building}

\end{figure}

The stereo estimation algorithm consists of four main steps, namely: feature extraction, matching cost calculation, disparity estimation and disparity refinement. While traditional methods, such as SGM~\cite{reg:sgm} or MGM~\cite{reg:mgm} use handcrafted functions and features for the stereo estimation algorithm, recent work has seen major improvements by exchanging one or all of the steps using deep learning. Deep-learning approaches, especially the use of convolutional neural networks, in short CNNs have been successfully applied to many areas of computer vision, often improving upon traditional methods in speed and performance. One of the biggest drawbacks of deep learning methods however, is the reliance on accurate ground-truth data. Creating such ground truth data is time consuming and expensive and therefore reliable ground-truth data sources are not available for many domains. Furthermore, the generality of networks trained on a specific domain is questionable. Previous works, such as the work from L. Hu et al.~\cite{sat_generality_segm} have shown that the generality of models is especially difficult for satellite images when trained with data from different continents. The creation of such ground truth for specific scenes is costly and often many hours of manual labour are needed for it. Especially creating detailed ground-truth data for the 3D-reconstruction task of urban scenes is often a futile task, as the scene will likely have changed by the time the data set is completed, as can be seen in Fig.~\ref{fig:missing_building}. In this example, taken from the DFC2019~\cite{data_fusion} dataset, a building can be seen in the image that is not present in the ground truth disparity data.

\begin{figure}[t]
  \centering
   \includegraphics[width=1.0\linewidth]{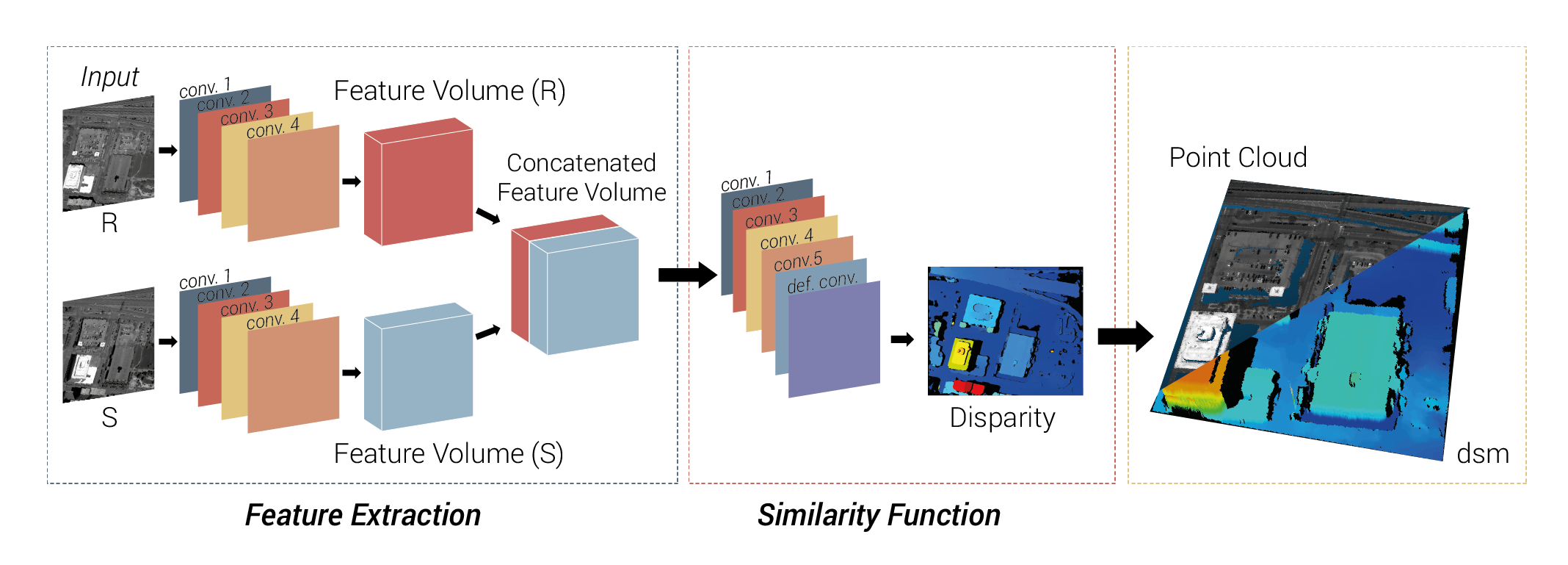}
   \label{fig:network}
   \caption{Overview of our method. The input consists of two stereo rectified satellite image tiles. Our method does not need any additional input in order to be trained. We use the satellite stereo pipeline software (s2p) in order to project our depth estimation into world coordinates and create a point cloud as well as a digital surface model (dsm).}
\end{figure}

We tackle this problem by creating a training routine that is completely independent from any such created ground truth and only uses the rectified panchromatic images as input in order to guide the training process. Other such self-supervised stereo methods use the photo-consistency loss for training by warping one image using the estimated disparity map to warp the other image and then calculating the similarity between the warped image and itself. Using this loss on its own however will lead to some artefacts, especially for homogeneous areas~\cite{monodepth}. Therefore, many works suggest to use a combined loss, for instance combining the photo-consistency loss with some sort of regularization term, like disparity smoothness or correspondence consistency between the prediction for the left and right image frame~\cite{monodepth}\cite{selfsup:monodepthv2}. Instead of the photo-consistency loss, we use a 'pseudo-ground truth' created by our initialization step. We use the left-right consistency check~\cite{lr-check} in order to remove inconsistent points from the disparity map and consider the remaining disparity values as already correct. This pseudo ground-truth is then consistently updated after each training step. The evolution of such a ground truth over the epochs can be seen in Fig.~\ref{fig:disp_evo}. Furthermore we show that the amount of inconsistent points correlates strongly with the amount of incorrect points and can therefore be used to track the training process of the network. We show that while some consistent points in the ground truth are wrong, it does not influence the overall accuracy of the training process. An overview of the whole method is illustrated in Fig.~\ref{fig:network}.



In summary, our contributions are as follows:
\begin{itemize}
    \item We create a novel training scheme based on the left-right consistency check that does not rely on any ground-truth data, making our training truly self-supervised.
    \item We create a CNN structure that is simple, lightweight and usable on most commercial hardware and yields good results for many different use-cases. With only $495K$ total trainable weights, it is lightweight when compared to other deep learning stereo methods which often use millions of trainable parameters.
    \item We show that our method produces good results on difficult real life scenes taken from the WorldView-3 satellite.
    
\end{itemize}

\section{Related Work}
Our work is based on previous works on stereo methods and self-supervised machine learning.

\textbf{Traditional stereo methods} use handcrafted features and similarity functions in order to match corresponding image points between two rectified image frames of the same scene. They can be grouped into three major groups, differing in their approach. \textbf{Local approaches} ~\cite{local1}~\cite{local2} are fast but in general lack the accuracy of other methods. \textbf{Global methods}~\cite{global1}~\cite{global2} on the other hand have high accuracy, however their computational complexity make them unsuitable for many real life tasks. \textbf{Semi-global} approaches are a good trade-off between accuracy and computational complexity and are therefore the most popular used methods. 

Semi-global matching (SGM)~\cite{reg:sgm} from H. Hirschmüller approximates a 2D smoothness constraint by using multiple 1D line optimizations for each pixel in order to refine the overall accuracy of the disparity estimation. 

G. Facciolo et al.~\cite{reg:mgm} improves on this principle, by creating and using different aggregation elements. Instead of using the 16 cardinal directions for cost propagation, such as described by the Semi-global matching method, he creates and uses more complex structures. This helps with the block artefacts that can be produced by the update scheme of Semi-global matching. Even with the advances of machine-learning based methods, More global matching (MGM) is still a viable method that produces good results for real life examples and is therefore still a popular method in the remote-sensing community.

\textbf{Deep learning stereo estimation} has been an successful endeavour in the last decade. One of the first deep learning based stereo methods is by J. Zbontar and Y. LeCun called MC-CNN~\cite{disp:mc_cnn}. In this work they popularized the shared-weight siamese network structure for stereo estimation. Variations of this structure has since been adapted by many state-of the art learning based stereo methods~\cite{disp:fcdcnn}\cite{disp:fcdsn_dc}\cite{disp:ga_net}\cite{disp:psm_net}\cite{disp:gc_net}\cite{raftstereo}.
In their work, J. Zbontar and Y. LeCun furthermore define the training task as a binary classification task, where matching image patches are defined as the positive classes and non-matching patches are defined as the negative classes. The training goal is then to maximize the similarity of positive class samples while minimizing the similarity of negative samples.

GC-Net~\cite{disp:gc_net} uses 3D convolutions in order to regularize the cost-volume and uses soft-argmin for subpixel accuracy. Ga-Net~\cite{disp:ga_net} improves upon the results of GC-Net by getting rid of many of the time- and memory-intensive 3D convolutions. Instead they introduce a cost-aggregation block that uses less 3D convolutions and consists of a semi-global and local aggregation part in order to refine the cost-volume.

In their work J.R. Chang et al. ~\cite{disp:psm_net} use a pyramid stereo matching network. They first use a pyramid pooling structure in order to improve context information for trained image features. Afterwards they create a cost-volume using this trained features. In the last step, the cost-volume is fed into stacked hourglass module using 3D-convolutions and a regression task is used for the final depth prediction.

J. Li et al. uses convolutional GRUs in a multi-level fashion in order to improve information propagation across the image. Their method, called RAFT-stereo~\cite{raftstereo} uses iterative refinement, traditionally used for optical flow estimation for the stereo task. They extract correlation features from the images at different resolutions and use the recurrency of the GRUs in order to iteratively improve the disparity estimation.

Following MC-CNN~\cite{disp:mc_cnn}, our method uses a variation of the shared-weight siamese network structure. We use the same definition of the training classes as MC-CNN. We reformulate the min-max problem as a hinge-loss function. This is the same formulation that has been used in FC-DCNN~\cite{disp:fcdcnn} and FCDSN-DC \cite{disp:fcdsn_dc}.

\textbf{Self-supervised machine learning} refers to the task of learning a model without having corresponding human-annotated labels. In recent years, this method of training has seen success on a wide area of computer vision tasks including depth estimation. 
MonoDepth by C. Godard et al.~\cite{selfsup:monodepth} is among the most popular self-supervised depth estimation networks. While in their case, inference can be done on a single image frame (Monocular Depth Estimation), the method is trained on two image frames (Binocular or Stereo Depth Estimation). C. Godard et al. improves upon this method with their second version called MonoDepthv2~\cite{selfsup:monodepthv2}. In particular they use an improved photo-consistency loss and introduce a mask that finds image points without any ego-motion. For the final prediction they use multi-scale images. With these additions they produce considerably better results. 

Ma F. et al. create a self-supervised Sparse-to-Dense deep learning approach for monocular depth prediction in their work~\cite{self-sup-sparse2dense}. They create a deep regression network that takes sparse 3D LiDAR (Light Detection and Ranging) points for supervision and use stereo color images and the photometric warping loss together with a smoothing loss in order to predict a dense depth map.

P. Knöbelreiter et al. in their work~\cite{cnn-crf-self} use their pre-trained CNN-CRF~\cite{cnn-crf} model in order to generate ground truth data for the Vaihingen dataset for 3D reconstruction~\cite{vaihingen}. They first create the disparity maps using their pre-trained model, then remove inconsistent points using the left-right consistency check. The so created disparity maps are then used in order to fine-tune their already trained network. This leads to an improvement of the reconstructed scene.

In this work, instead of a warping or photo-consistency loss we introduce an adaptive pseudo ground-truth loss. We directly predict occluded points (i.e. points without ego-motion) by using the left-right consistency check~\cite{lr-check} and do not rely on any additional input data. 



\section{Remote-sensing Stereo Matching}
In this work we use the area of interest (AOI) from the 2019 IEEE Data Fusion Contest (DFC2019)~\cite{data_fusion}. Furthermore we use the satellite stereo pipeline s2p~\cite{s2p} for the tiling, rectification and projection step.

\begin{figure}[t]
    \begin{center}
   \includegraphics[width=0.4\linewidth]{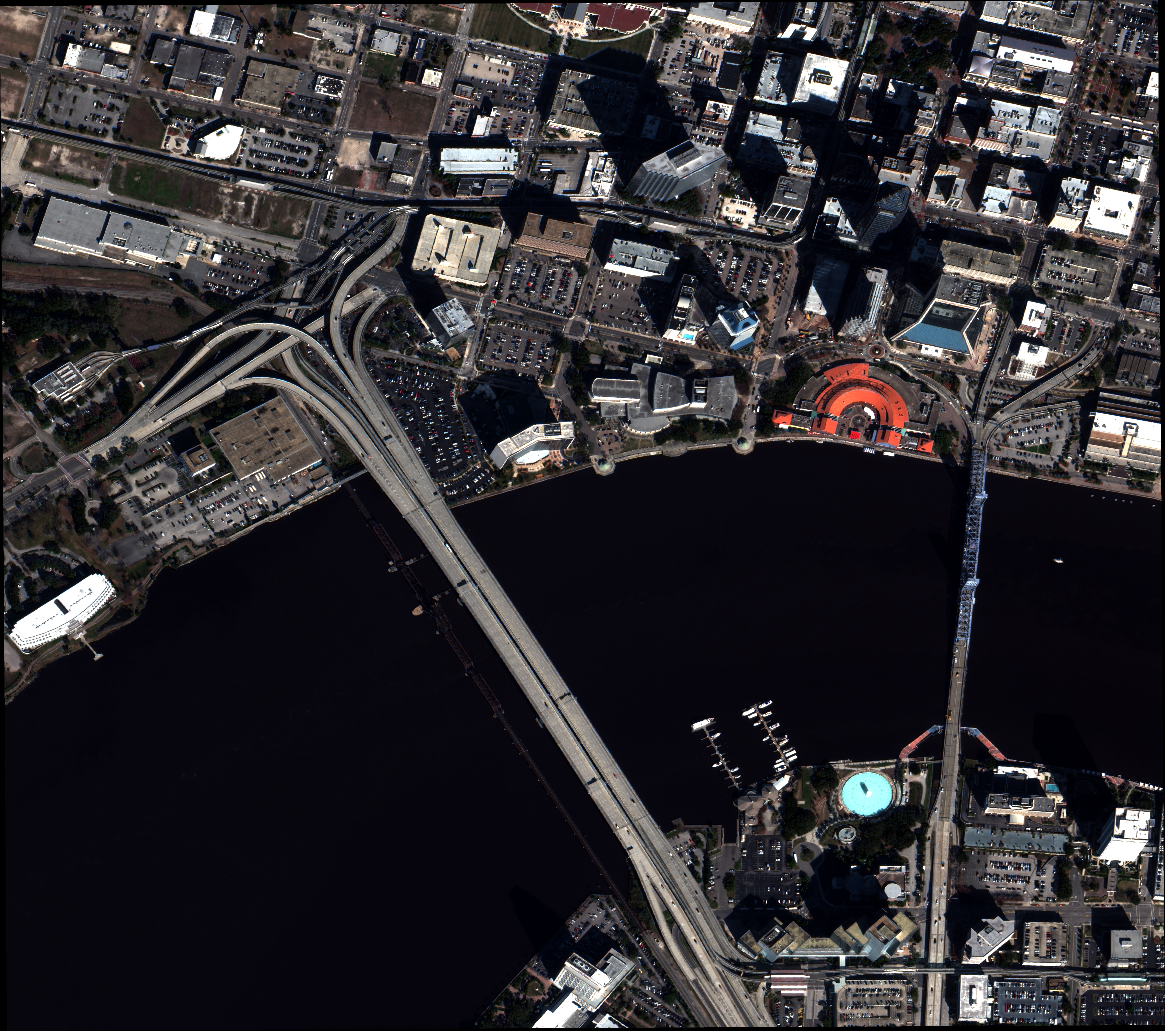}
   \end{center}
   \includegraphics[trim={0 0 0 26px},clip=true,width=0.3\linewidth]{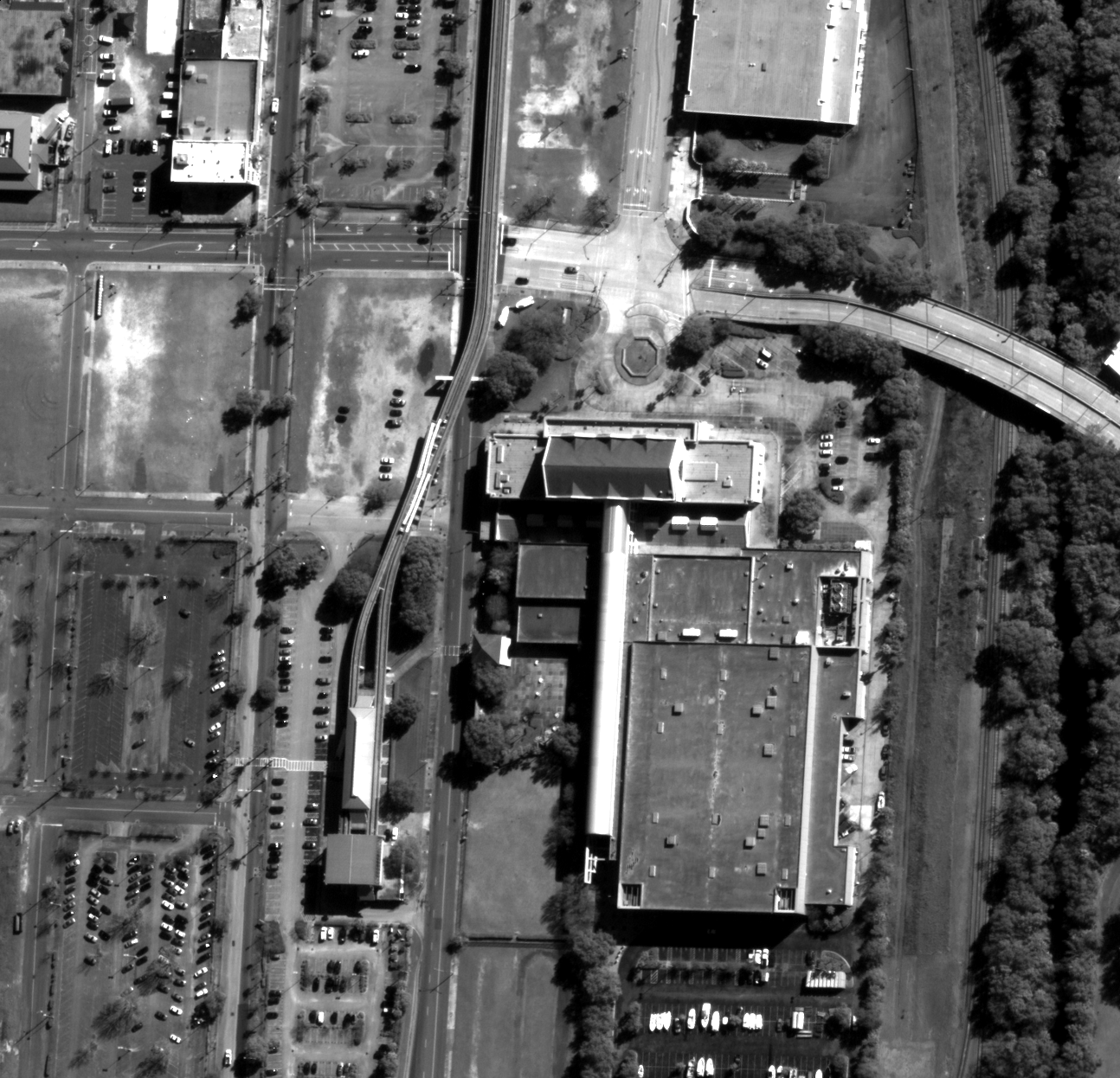}
   \includegraphics[width=0.3\linewidth]{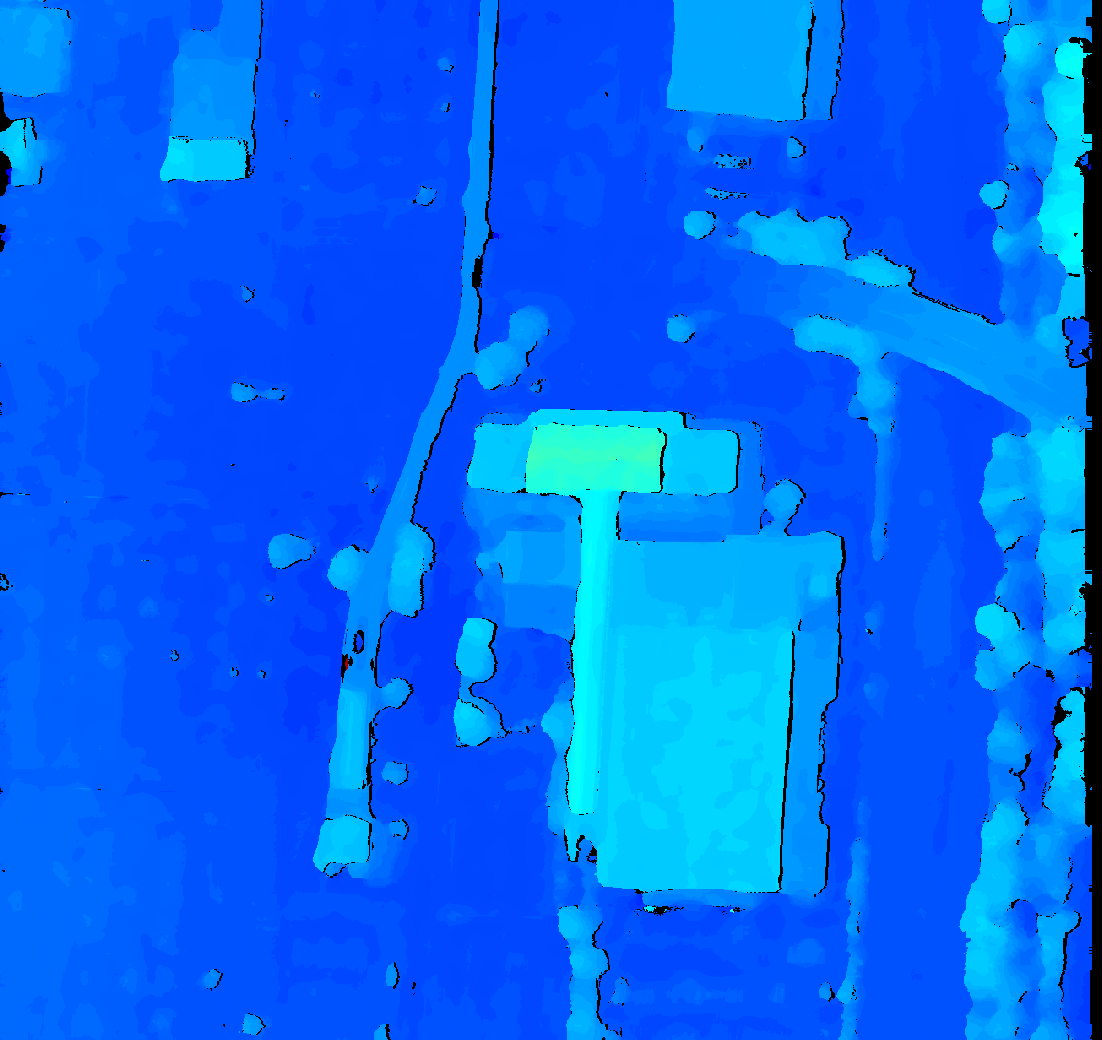}
   \includegraphics[trim={0 0 0 13px},clip=true,width=0.33\linewidth]{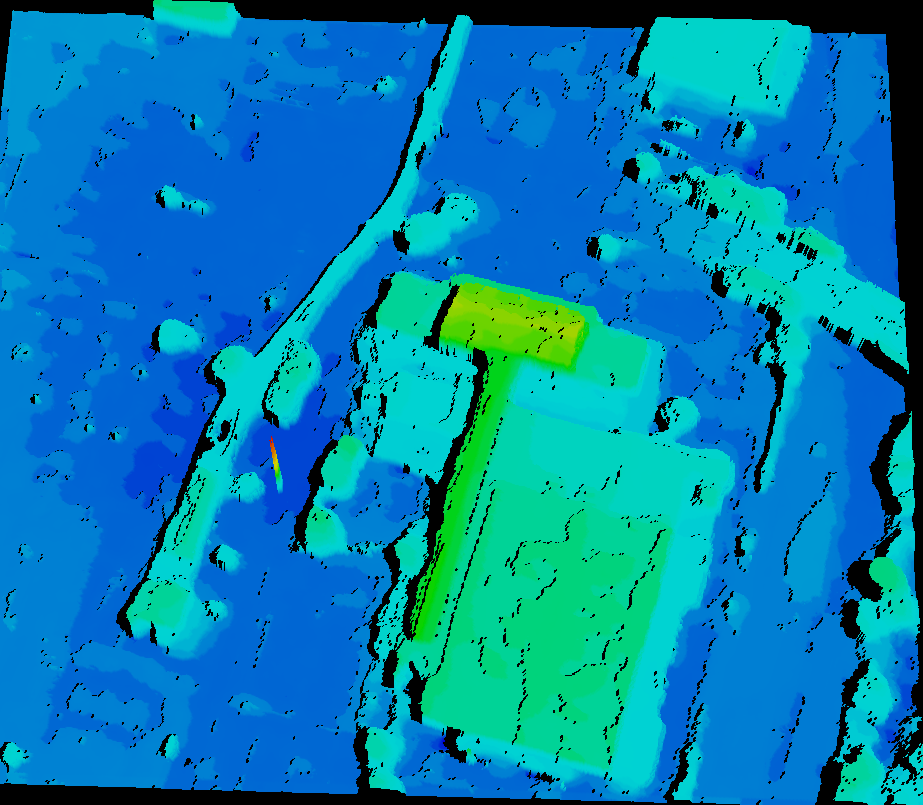}
   \caption{First row: AOI from Jacksonville, Florida USA using the RGB bands of the multispectral image instead of the panchromatic image for the sake of better visualization. This image was captured using the WorldView-3~\cite{wv3} satellite in january 2015. Second row f.l.t.r: one reference tile of the area of interest, the predicted disparity map and the projected digital surface model that resulted from it.}
   \label{jack}
\end{figure}

This dataset also provides a digital surface model (dsm) for the AOI, which was created using a LiDAR scan. We use this provided dsm for metric evaluation in our work. For the evaluations, the metrics as described by M.Bosch in their work ~\cite{metrics} is used. 
An RGB image of the area of interest used in this evaluation, as well as an example of an extracted rectified reference tile, the disparity map predicted  by our method and the resulting projected dsm can be seen in Fig.~\ref{jack}.

For this area, a ground truth dataset for the stereo matching task has been created. Rectified image pairs together with corresponding disparities have been released to the public. 
However, we only use the ground truth for the calculation of the end-point error, not for training itself. For the training, we use satellite images from the SpaceNet challenge dataset~\cite{spacenet} of the same area of interest. To this end we use the panchromatic, not the multispectral images provided by the WorldView-3 satellite~\cite{wv3} as this band has a ground resolution of 0.31m per pixel while the eight-band multispectral imagery has a ground resolution of 1.24m per pixel.

\section{Training Loop and Network}


\begin{figure}[t]
  \centering
\includegraphics[width=0.99\linewidth]{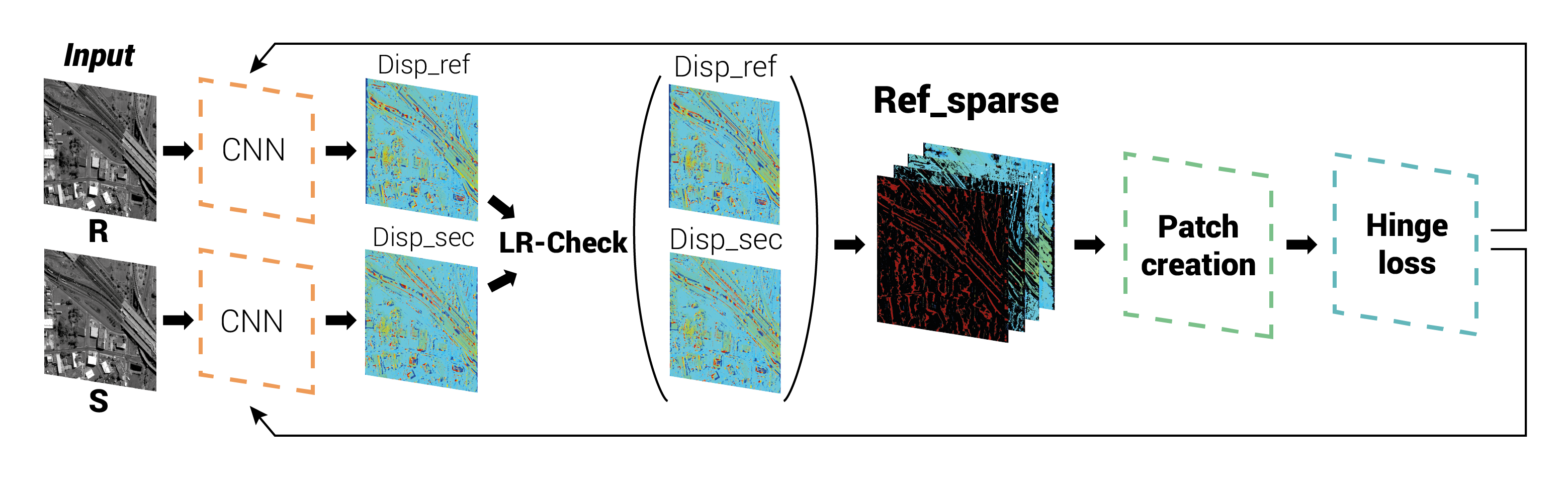}
   \caption{Schematic of our training loop.}
   \label{fig:train1}
\end{figure}

\begin{figure}[t]
  \centering
   \includegraphics[width=0.99\linewidth]{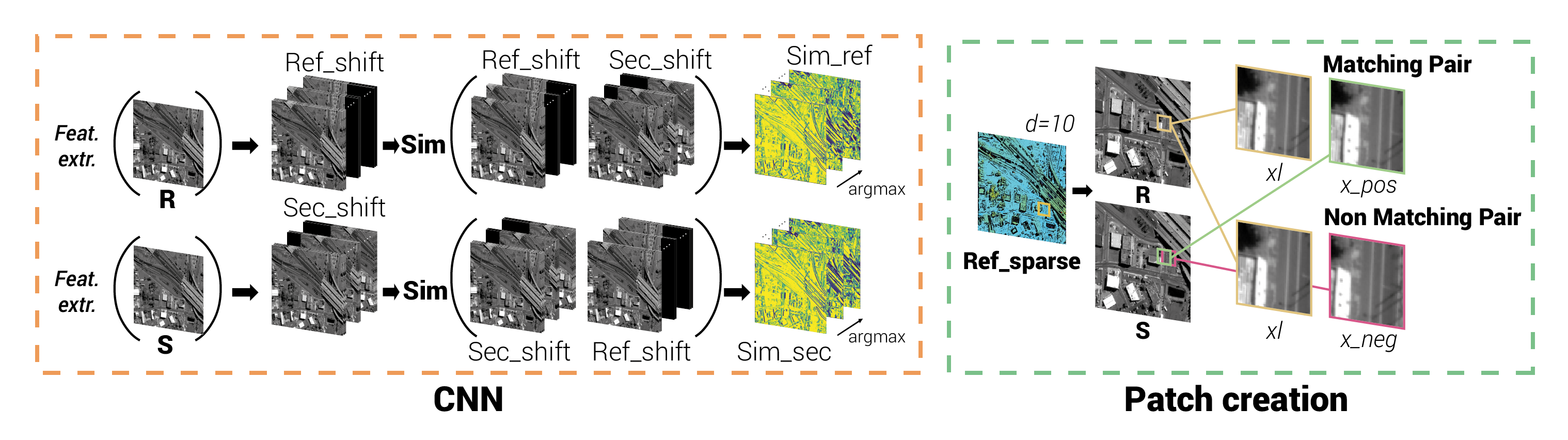}
   \caption{Detailed illustration of the CNN and patch creation part of our training loop.}
   \label{fig:train2}
\end{figure}

Fig.~\ref{fig:train1} is a visual representation of how the training loop of our method works. First, two stereo rectified satellite images are put into our CNN which creates the disparity estimation for each frame. In the next step, the left-right consistency check~\cite{lr-check} is used in order to get rid of inconsistent points which creates the sparse disparity map. This sparse disparity map is then used for the training patch creation for the current iteration. Using this patch, the hinge-loss is calculated and then back-propagated through the network. Fig.~\ref{fig:train2} shows more detailed illustrations of our network (orange dashed box) and how the patches used for training are created (green dashed box).

Our network consists of two main parts. The first part learns rich and deep image features of the satellite images, the second trains similarity functions between extracted features, improving upon handcrafted similarity functions. 

The output of the feature extraction network part is a 60-dimensional image, e.g. $H\times W\times 60$. For the sake of interpretability, we continue to show the panchromatic image in Fig.~\ref{fig:train2}. The trained deep feature images of the reference and the second image are concatenated and used as input for the similarity function part, the output is a 1 dimensional similarity measurement e.g. $H\times W\times 1$.

The green dashed box shows how the sparse disparity map of the current iteration is used for training patch creation. First, a patch around a random point with a consistent disparity value (in the current sparse ground truth) is chosen. The corresponding patch (in the same location) from the panchromatic reference image is extracted. Then the disparity value is used in order to get the position of the matching patch in the second image. A random small offset along the horizontal axis is then added for the non matching crop of the second image. After each training epoch, the sparse disparity maps are updated again and used as pseudo ground truth for the next epoch.

\subsection{Implementation Details} 

We implement our method using Python3, pytorch 1.8.0~\cite{pytorch} and Cuda 12.4. We use gdal 2.4.2 for the manipulation of geo-tiffs and OpenCV~\cite{opencv} for other image manipulation. A single NVIDIA GeForce RTX 3090 consumer grade GPU is used for training. The network is trained using the Adam optimizer~\cite{adam} with a learning rate of $6.0 \times 10^{-5} $. We use randomly cropped patches from the reference and second image with a size of $11x11$ and a total batch-size of $500$ for training.
They are trained using a hinge-loss which maximizes the similarity between matching image patches and minimizes the similarity between close non-matching patches. Let $s_{+}$ be the similarity between two matching image patches extracted from the reference and the second image and $s_{-}$ be the similarity between two non-matching image patches, then the loss is defined as a hinge-loss, as seen in Eq.~\ref{eq:sim_loss}.

\begin{equation}
    loss = max(0, 0.2 + s_{-} - s_{+}).
    \label{eq:sim_loss}
\end{equation}

Following previous works called FCDSN-DC~\cite{disp:fcdsn_dc} abd FC-DCNN~\cite{disp:fcdcnn} this loss is implemented using ReLU~\cite{act:relu}, which leads to a slight reformulation. 
The adapted loss can be seen in Eq.~\ref{eq:reform}.
\begin{equation}
    loss = ReLU(s_{+} - s_{-} - 0.2).
    \label{eq:reform}
\end{equation}

\subsection{Pseudo Ground-Truth}
Our self-supervised training is based on an adaptive sparse-to-dense update scheme. For the initial step, the disparity map is calculated using the panchromatic image features from the satellite image with the cosine similarity function. Then, the left-right consistency check~\cite{lr-check} is used in order to remove inconsistent points. The left-right consistency check is defined in Eq.~\ref{eq:lrcheck}.

\begin{figure}[t]
  \centering
   \includegraphics[width=0.2\linewidth]{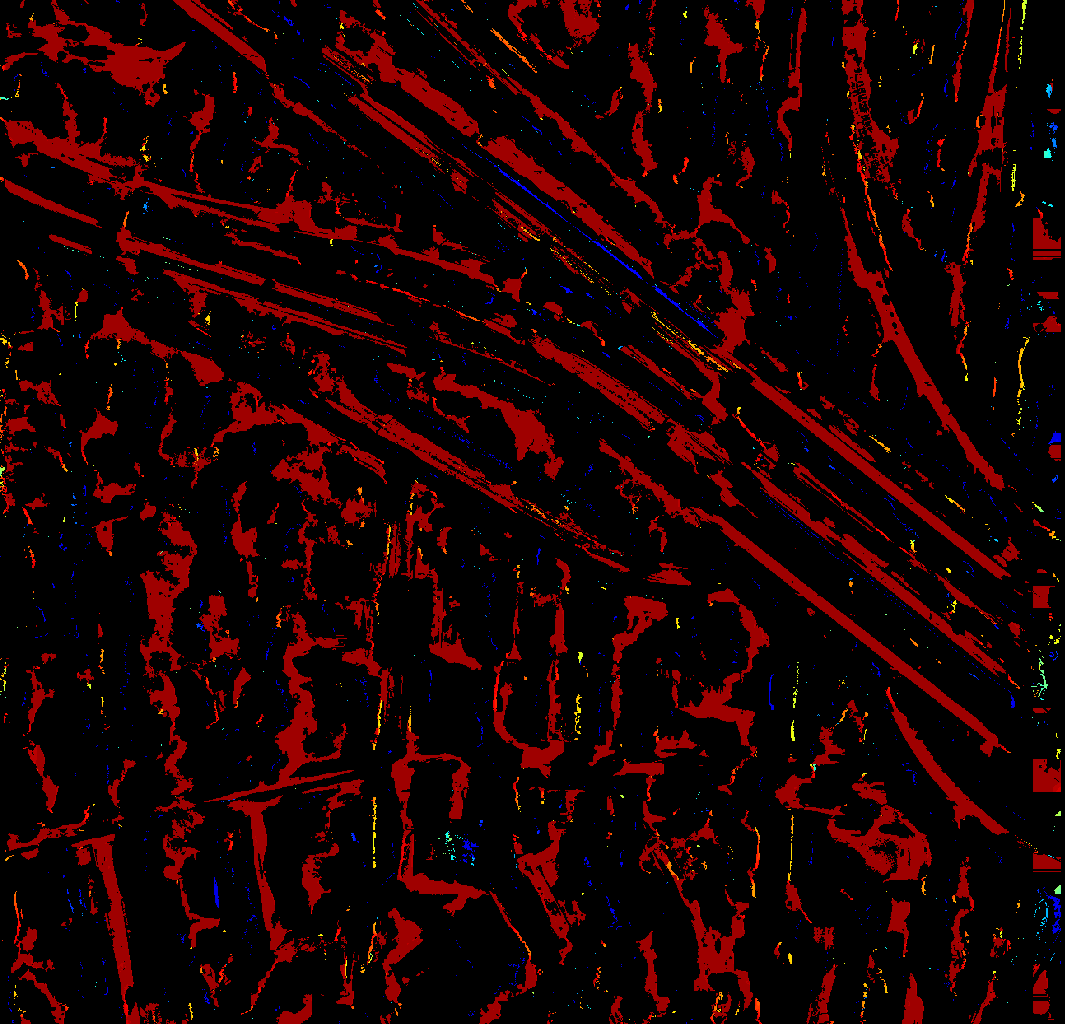}
   \includegraphics[width=0.2\linewidth]{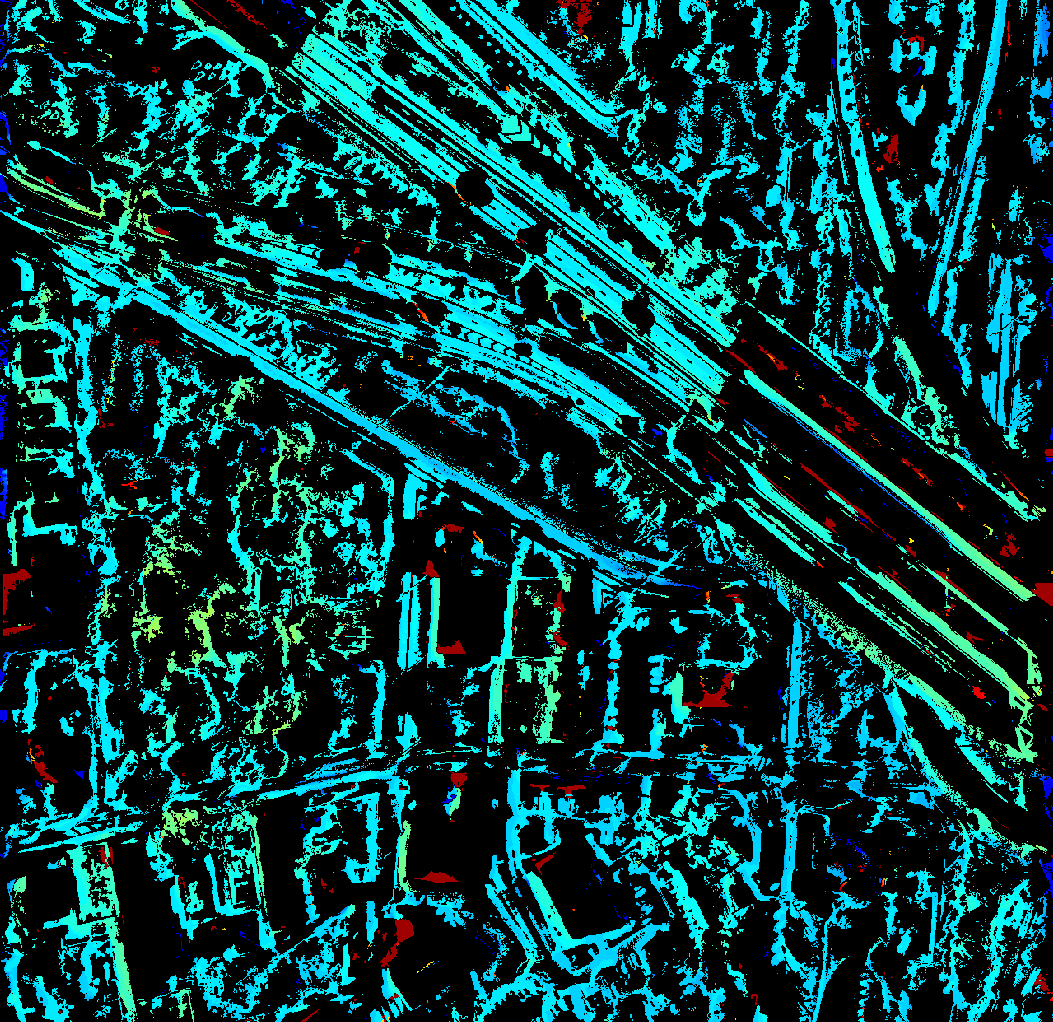}
   \includegraphics[width=0.2\linewidth]{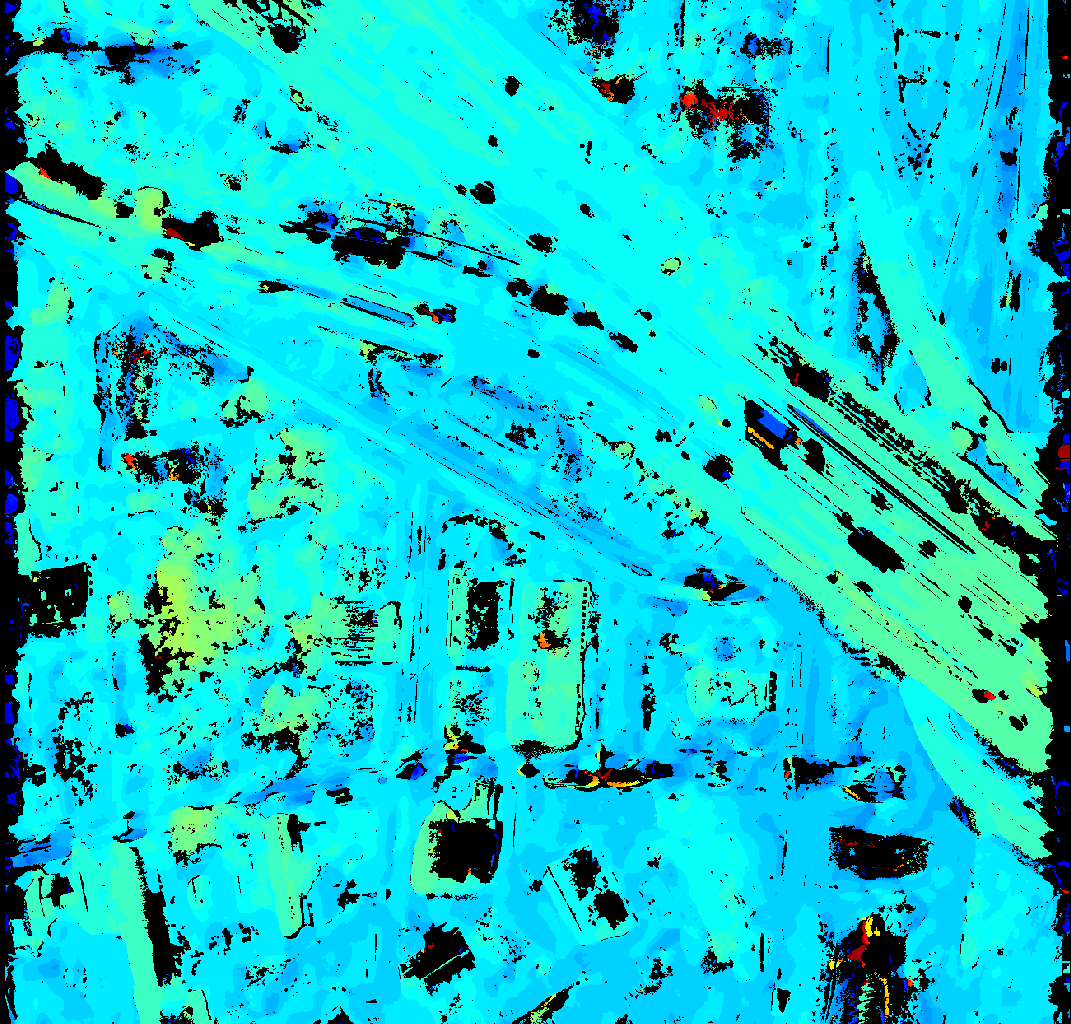}
   \includegraphics[width=0.2\linewidth]{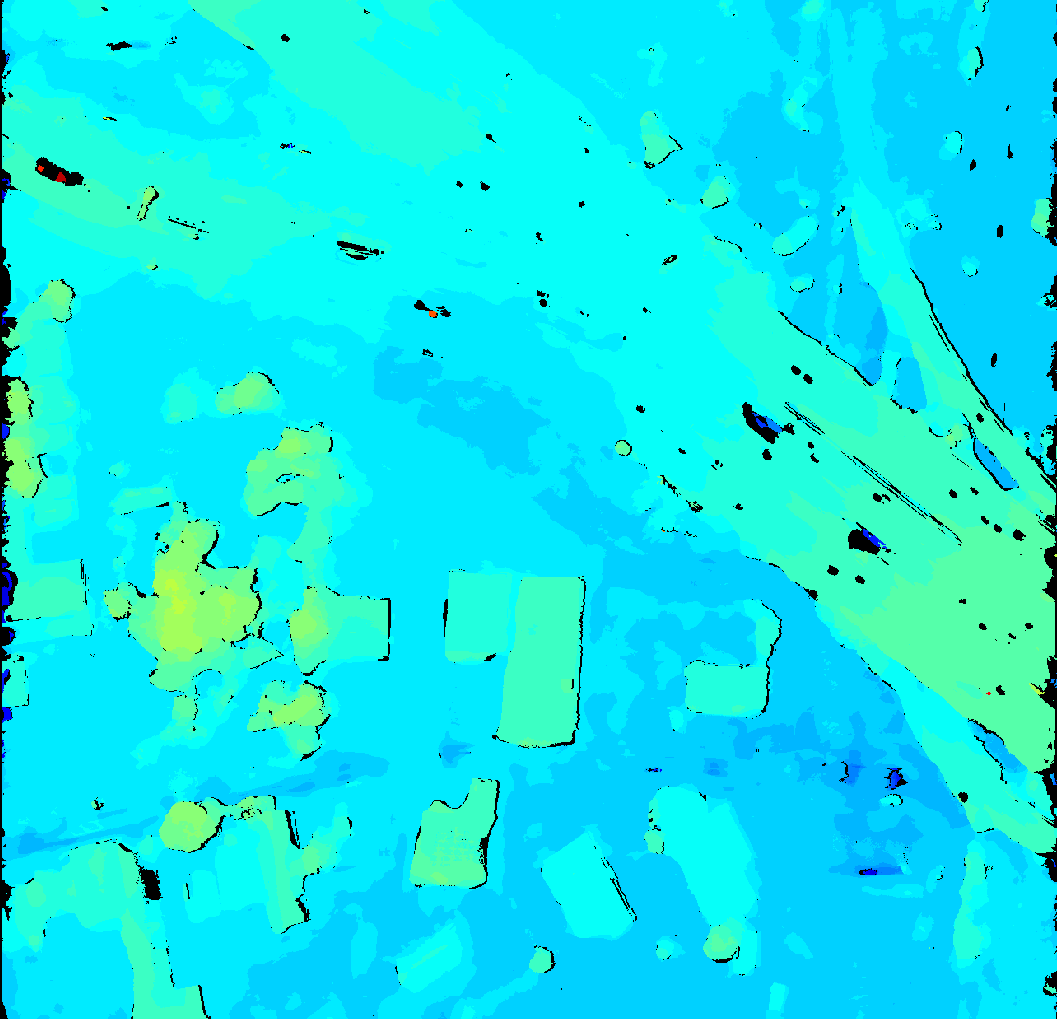}
   \includegraphics[width=0.8\linewidth]{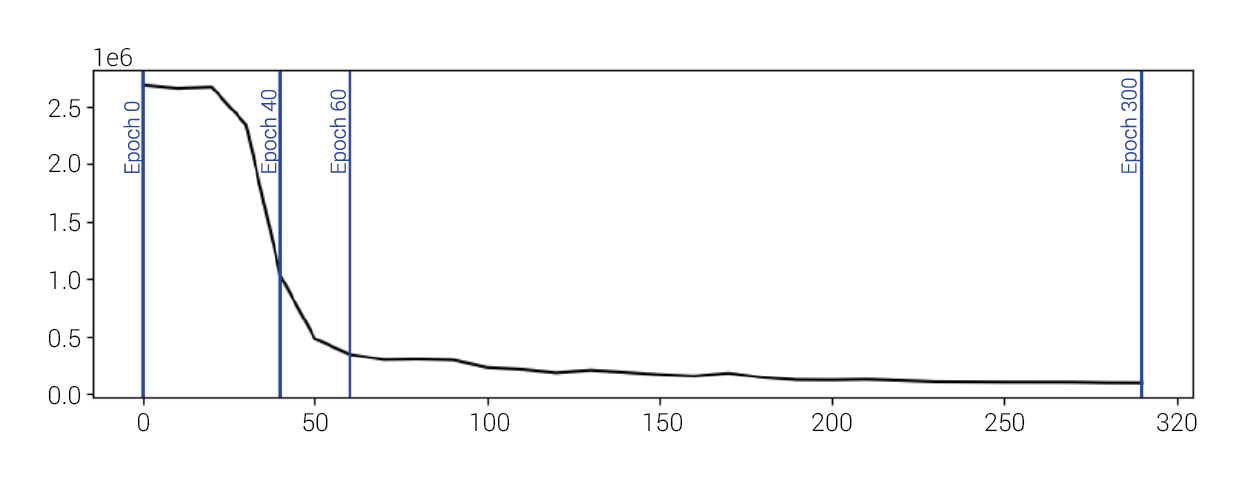}
   
   \caption{Evolution of the disparity map used as ground truth for self-supervision. F.l.t.r.: Epoch 0, Epoch 40, Epoch 60 and Epoch 300. The evolution of the number of inconsistent points can be seen in the second row plot. The blue lines indicate the shown disparity maps.}
   \label{fig:disp_evo}

\end{figure}

\begin{equation}
   |D^{L}(x,y) - D^{R}(x-d,y)| > 1.1.
   \label{eq:lrcheck}
\end{equation}

This evolution of the predicted disparity map can be seen in Fig.~\ref{fig:disp_evo}. It shows that even if you start with a very sparse disparity map as initial ground truth, our update scheme produces dense and accurate maps after only 300 training epochs. Furthermore, it shows that in early epochs of the training process, only easy to match image points, such as corners or edges of foreground objects (i.e. roads and buildings) are consistent. Using this strongly matching pixels as a starting point, the network slowly learns more ambiguous image features. In theory, if no strong matching image points are found in the initial disparity map, the network could get stuck early during training, however this has never happened in practise during our experiments. 
The second row of Fig.~\ref{fig:disp_evo} shows the total number of inconsistent points used for the tracking. In our experiments, if the number of inconsistent points increases for 50 consecutive epochs, we stop the training process.

We argue, that the total sum of such removed points correlates with the amount of incorrect points and this can therefore be used for tracking network convergence and early-stopping. 
In order to show this, we conduct an experiment on 20 randomly chosen image pairs for the training split and 20 randomly chosen image pairs for the test split from the 2019 IEEE Data Fusion Contest 2019 (DFC2019)~\cite{data_fusion} challenge. We use the end-point error, as defined in Eq.~\ref{eq:epe} in order to track the accuracy of the disparity map. The threshold $\tau$ gives the number of how close the prediction has to be to the ground truth in order to be counted as correct. For example when using the 4-point error ($\tau = 4$), every predicted point which is within the range of $\pm 4$ to the ground truth is counted as being a correctly predicted image point. 

\begin{equation}
    \sum |D_{gt} - D_{pred}| >= \tau .
    \label{eq:epe}
\end{equation}

\subsection{Sub-Pixel Enhancement}

We follow the work of V.C. Miclea et al.~\cite{subpx} for our sub-pixel enhancement scheme. While many methods use machine learning to learn sub-pixel residuals for the disparity estimation, the feasibility and accuracy of those methods in a self-supervised training framework is an open research question to the best of our knowledge. The algorithm we use in our method is defined as follows: 

  \begin{equation*}
    d_{subpx} =
    \begin{cases}
      d_{Int} - 0.5 + arctan(\frac{ld}{rd}), & \text{if}\ ld \leq rd \\
      d_{Int} - 0.5 + arctan(\frac{rd}{ld}), & \text{otherwise}
    \end{cases}
  \end{equation*}
  
  \begin{equation}
    ld = c_{d-1} - c_{d}.
    \label{eq:subpx}
  \end{equation}
  
  \begin{equation*}
    rd = c_{d+1} - c_{d}.
  \end{equation*}

Here, $d_{Int}$ is the chosen integer disparity value for the image point we want to get the sub-pixel value for. Depending on the implementation, this is either the position with the highest or lowest value in the cost volume. This cost (or similarity) is defined as $c_{d}$. Following $c_{d-1}$ is the cost of the next pixel to the left of the highest matching pixel and $c_{d+1}$ is the next pixel to the right of the most similar image point in the second image. In our implementation, the sub-pixel enhancement is used on the consistent points not removed by the left-right consistency check. It is not used in order to find more inconsistent points in the disparity map.


\section{Experiments}

In this section, we compare the scores produced by our method with the scores of other popular deep learning based methods and the baseline method used by the s2p~\cite{s2p} software MGM~\cite{reg:mgm}. 
We use the publicly available trained weights for the other deep learning methods. We do not retrain or fine-tune the well-known state-of-the-art methods, because we want to show their accuracy if there is no possibility of fine-tuning or training from scratch because no ground truth data exists for the specific scene. Even though ground truth depth data is available for this particular AOI, it is missing for most remote-sensing data. 

\begin{table}[ht!]
\center
\caption{Comparison on Jacksonville data}
\label{tab:jack_comp}
\begin{tabular}{|c|c|c|c|c|}
\hline
\hline
Method & MGM~\cite{reg:mgm} & SAdaNet (ours) & PSMNet~\cite{disp:psm_net} & RaftStereo~\cite{raftstereo} \\
\hline
recall & 0.894 & \textbf{0.906} & 0.800 & 0.828 \\
\hline
precision  & 0.843 & 0.836 & \textbf{0.891} & 0.829\\
\hline
jaccardIndex & 0.767 & \textbf{0.769} & 0.731 &  0.707\\
\hline
f-score & 0.868 & \textbf{0.870} & 0.8445 &  0.828 \\
\hline
\end{tabular}
\end{table}

We extend the s2p~\cite{s2p} pipeline, where the default matching algorithm is switched with our method or other deep learning based matching methods respectively for the experiment shown in Tab.\ref{tab:jack_comp}.

\begin{figure}[t]
  \centering
   \includegraphics[trim={0 1cm 0 0},clip=true,width=0.7\linewidth]{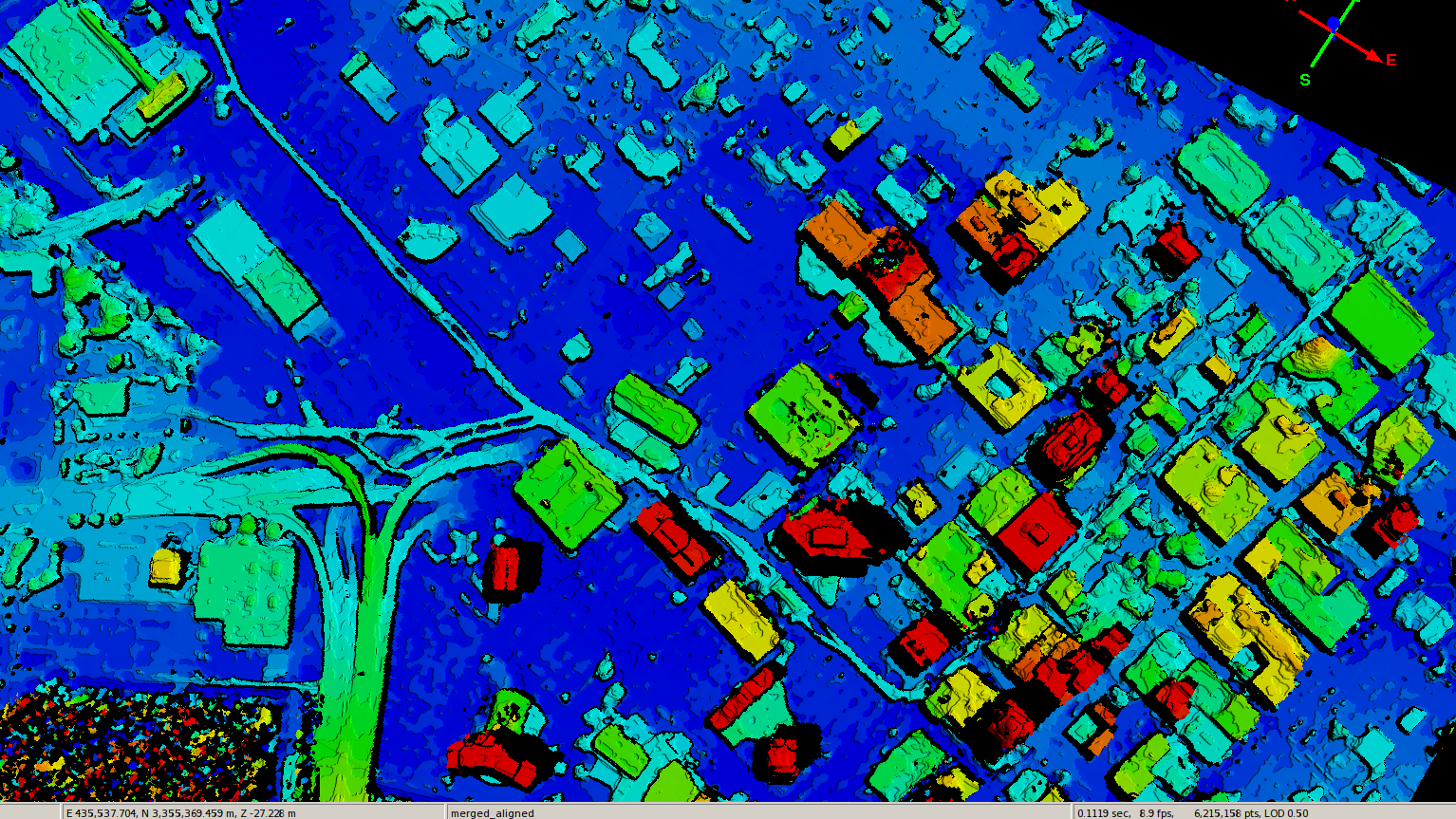}
   \caption{Part of the digital surface model of the evaluated area created by our method.}
   \label{fig:dsm_jack}
\end{figure}

As Tab.~\ref{tab:jack_comp} shows our self-supervised method outperforms all other evaluated methods in the evaluated area of interest with the exception of precision, where PSMNet is slightly better. However PSMNet produces a less complete digital surface model, which can be seen by the lower completeness and f-score. Therefore we argue that our method still compares favourably.
A result of our method, showing parts of the digital surface model of Jacksonville can be seen in Fig.~\ref{fig:dsm_jack}. 



\subsection{Ablation Study}

In this section we perform some ablation studies in order to show the validity and impact of our method. We perform every experiment on the same real life scenes taken from the WorldView-3 satellite as before, namely an area in Jacksonville, Florida USA.
First, we show our method only using the feature extractor part of the network with cosine similarity and no sub-pixel enhancement. Then, we use the feature extractor as well as the trained similarity part and no sub-pixel enhancement. The last step then shows the accuracy of feature extractor, trained similarity and sub-pixel enhancement together.

\begin{table}[ht!]
\center
\caption{Ablation study on Jacksonville data}
\label{tab:ablation_study}
\begin{tabular}{|c|c|c|c|}
\hline
\hline
Method & cosine similarity & trained similarity & \makecell{trained similarity +\\ sub-pixel enhancement} \\
\hline
recall & 0.896 & 0.902 &  \textbf{0.906}\\
\hline
precision  & \textbf{0.837} & 0.834 & 0.836\\
\hline
jaccardIndex & 0.763 & 0.765 & \textbf{0.769}\\
\hline
f-score & 0.865 & 0.867 & \textbf{0.870}\\
\hline
\end{tabular}
\end{table}

\begin{figure}[t]
  \centering
   \includegraphics[trim={0 0 0 0.9cm},clip=true,width=0.40\linewidth]{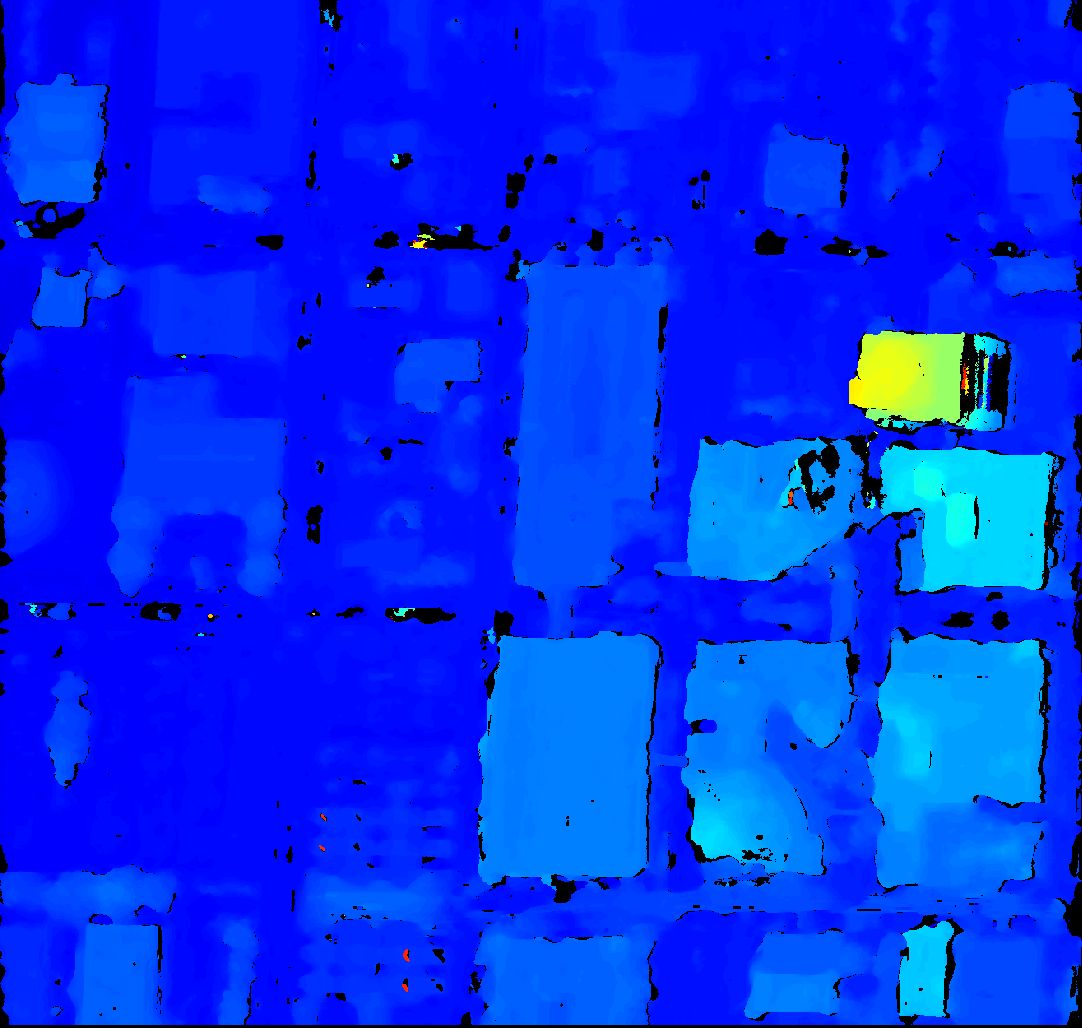}
   \includegraphics[trim={0 0 0 0.55cm},clip=true,width=0.40\linewidth]{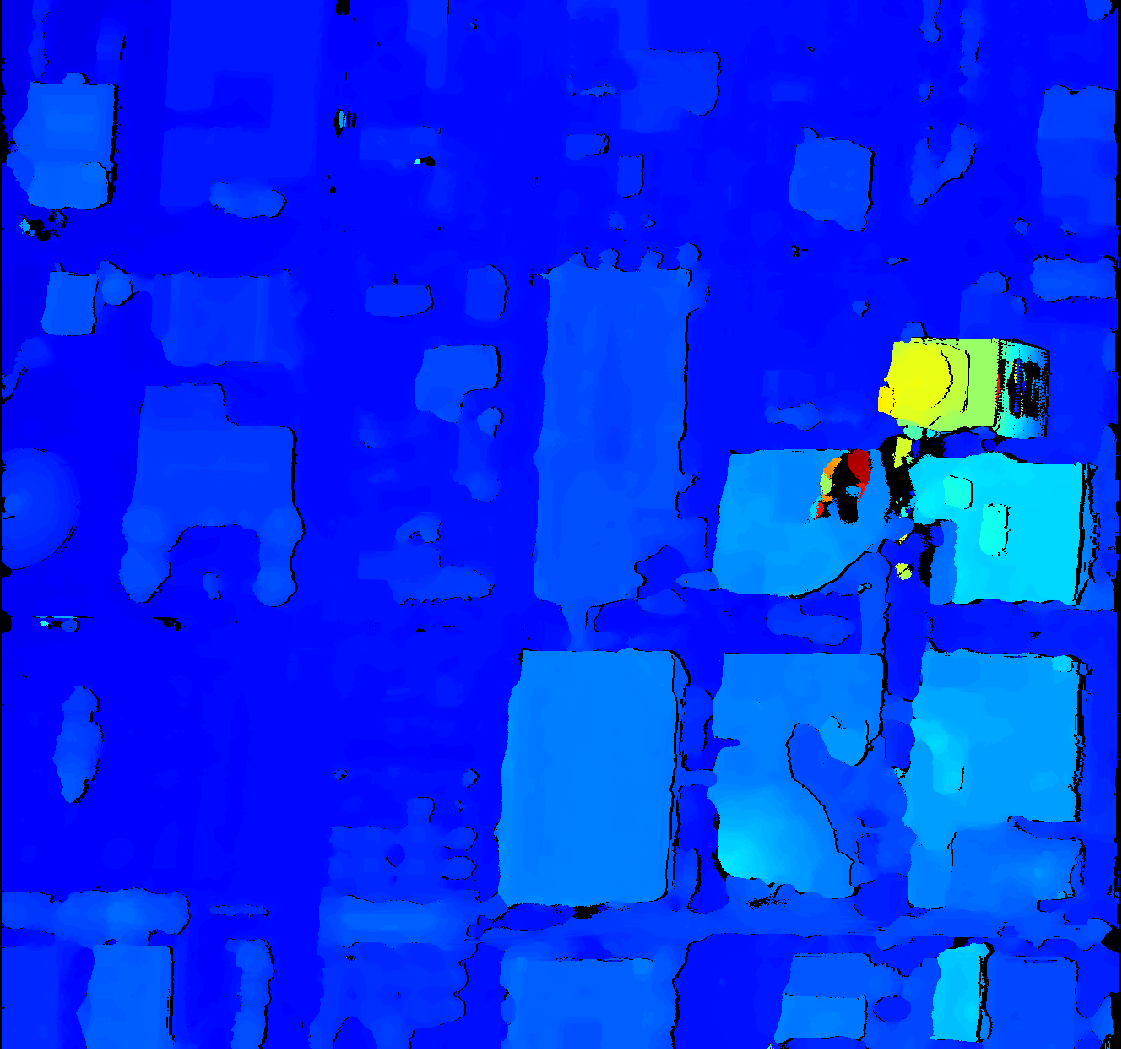}
   \caption{Left: Disparity map created using only the feature extractor network. Right: Disparity map created using both the trained feature extractor and the trained similarity function. Missing predictions are marked in black.}
   \label{fig:branch-simb}
\end{figure}

Tab.~\ref{tab:ablation_study} shows that the scores improve with each step, with the exception of the precision. This is expected as only using the trained feature extraction part of our method also leads to the sparsest reconstruction out of all three experiments conducted. As the whole area of interest is rather large, it is more difficult to show small differences between the methods. Therefore, to better illustrate the difference between only using the feature extractor and using the feature extractor plus trained similarity function, we show the disparity map of a small section from this ablation study created by these two methods in Fig.~\ref{fig:branch-simb}. Here one can see, that only training the feature extractor leads to a less complete reconstruction (inconsistent points are illustrated in black) when compared to training both the feature extractor as well as the similarity function.

\subsection{Point Error Analysis}

To further motivate the impact of our modules we conduct another experiment using the end-point metric and completion score in percentage from the available ground truth disparity from the DFC2019 dataset ~\cite{data_fusion}. Instead of projecting our prediction into 3D space and comparing it to the ground-truth digital surface model, where each point which is within 1m of the ground truth is counted as correct, we use the 4-point error and the 2-point error as to compare the 2D disparity maps for accuracy. For this $\tau$ in Eq.~\ref{eq:epe} is set to $4$ and $2$ respectively. We furthermore omit the sub-pixel enhancement for the evaluation, as we do not use it to detect additional inconsistent points, instead using it to refine consistent points in the disparity map.
For this evaluation we select 20 random images from the dataset and remove samples with structures missing in the ground truth, such as Fig.~\ref{fig:missing_building} or interference created by passing planes. Due to the difficulty of creating ground truth disparities for aerial imagery, the ground truth disparity of flat surfaces in the disparity map observed in the data set can vary by up to four pixels. Therefore we argue that the 4-point error is the best measurement for this specific dataset. However as the sub-pixel enhancement impacts lower threshold end-point errors more, we also report on the 2-point error in this experiment.
\begin{table}[ht!]
\center
\caption{Ablation study on DFC2019 data}
\label{tab:ablation_study_epe}
\begin{tabular}{|c|c|c|c|}
\hline
\hline
Method & cosine similarity & trained similarity & \makecell{trained similarity +\\ sub-pixel enhancement} \\
\hline
\multicolumn{4}{|c|}{Train}\\
\hline
4-PE & 12.064  & 8.979 & \textbf{8.514} \\
 \hline
2-PE & 27.376 & 24.124 & \textbf{18.079}\\
\hline
completion & 85.755 & \textbf{90.245}& - \\
\hline
\multicolumn{4}{|c|}{Test}\\
 \hline
 4-PE & 15.692  & 13.399 & \textbf{13.396} \\
\hline
2-PE & 32.753 & 30.191 & \textbf{24.557}\\
\hline
completion & 82.748& \textbf{84.603} & -\\
\hline

\end{tabular}
\end{table}

The first column of Tab.~\ref{tab:ablation_study_epe} shows the accuracy of our trained feature extractor with the cosine similarity function. The second column shows the improvement of the accuracy when the similarity function is trained as well. The last column shows the improvement in accuracy when the sub-pixel enhancement as defined in Eq.~\ref{eq:subpx} is used on the consistent points of the previous experiment.

\begin{figure}[t]
  \centering
   \includegraphics[width=0.9\linewidth]{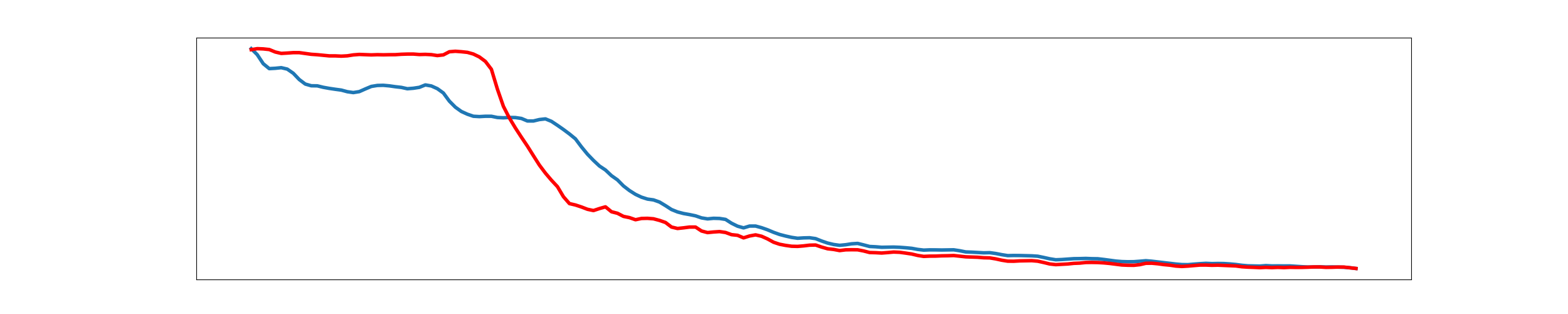}
   \includegraphics[width=0.9\linewidth]{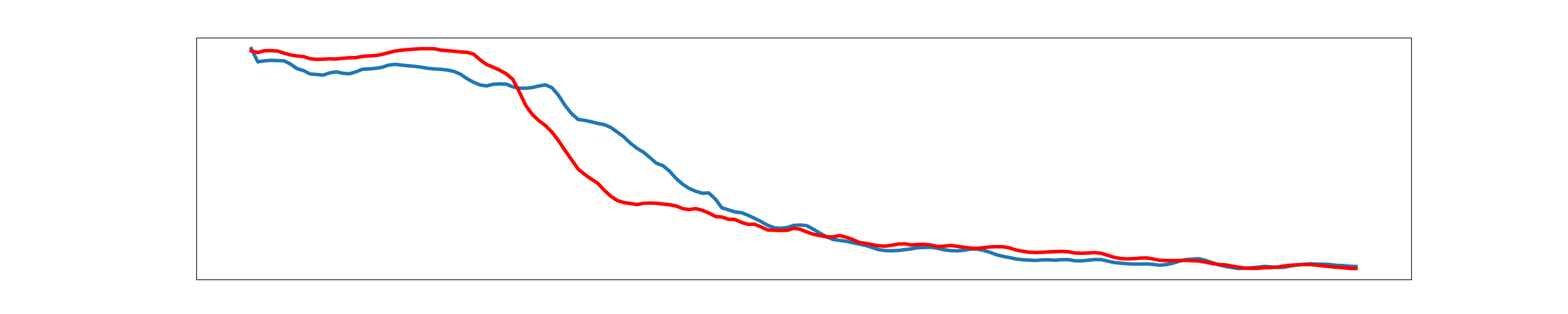}
   \caption{Evolution of the number of inconsistent points (blue) and the end-point error of the training split (first row) and test split (second row).}
   \label{fig:epeIncons}
\end{figure}

Tab.~\ref{tab:ablation_study_epe} shows the improvement in accuracy for each added step of our method for the train split as well as the test split. One can see that the accuracy of the method is stable for untrained samples, only sacrificing slightly on accuracy and completeness. Furthermore it shows that the sub-pixel enhancement improves the accuracy of higher threshold end-point errors, such as the 4-point error, only slightly. However it has an impact on the accuracy of lower threshold end-point errors, such as the 2-point error. In order to motivate the correlation between the number of inconsistent points with the number of wrongly predicted points, Fig.~\ref{fig:epeIncons} shows the evolution of both for the train and test split of this experiment. The blue line shows the total amount of inconsistent points which are removed using the left-right consistency check of the predicted output of the network after each training step. The red line shows the total amount of wrongly predicted image points after each training step. Fig.~\ref{fig:epeIncons} visualizes the correlation between the two metrics and that they converge to the same local minimum.

\subsection{Additional Datasets}

In order to show the generality and robustness of our method, we perform additional experiments on datasets from different domains.
To this end, we use the same hyperparameters, stopping criteria and training setup as for the previous experiments. We show qualitative and quantitative results for this experiments. As we want to cover a broad range of different domains, we train and evaluate our method on a dataset with indoor scenes, namely Middlebury~\cite{mb} as well as a dataset for autonomous driving called KITTI2015~\cite{kitti}.

\begin{table}[ht!]
\center
\caption{Results of our method on datasets from different domains}
\label{tab:epe_mb_kitti}
\begin{tabular}{|c|c|c|c|c|}
\hline
Dataset & 4-PE & 3-PE & 2-PE & 1-PE\\
\hline
Middlebury & 18.054 &  19.820 & 22.526 & 29.196\\
 \hline
KITTI2015 & 5.613 & 8.260 & 16.337 &  41.51\\
\hline

\end{tabular}
\end{table}

\begin{figure}[t]
  \centering
   \includegraphics[width=0.45\linewidth]{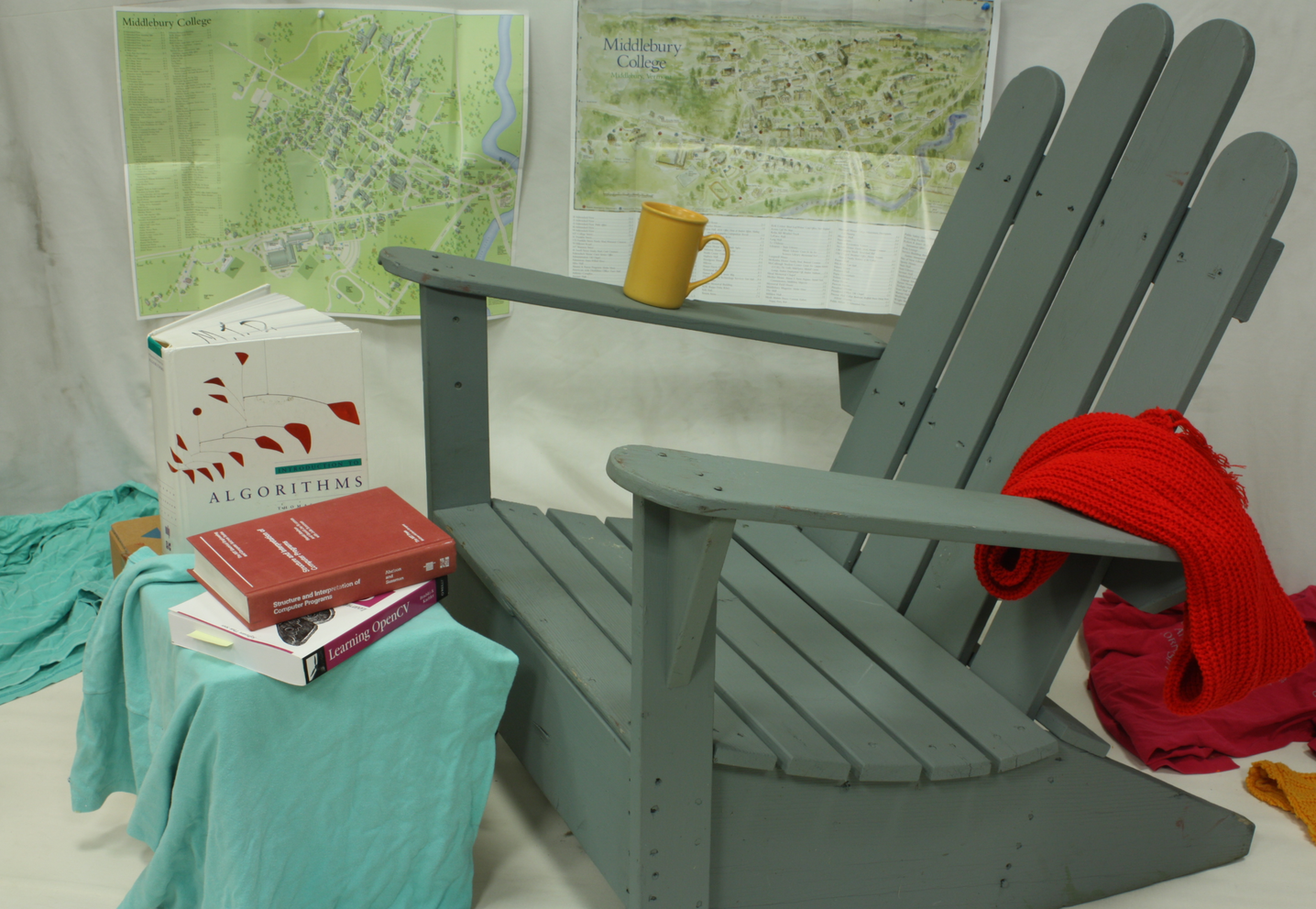}
   \includegraphics[width=0.45\linewidth]{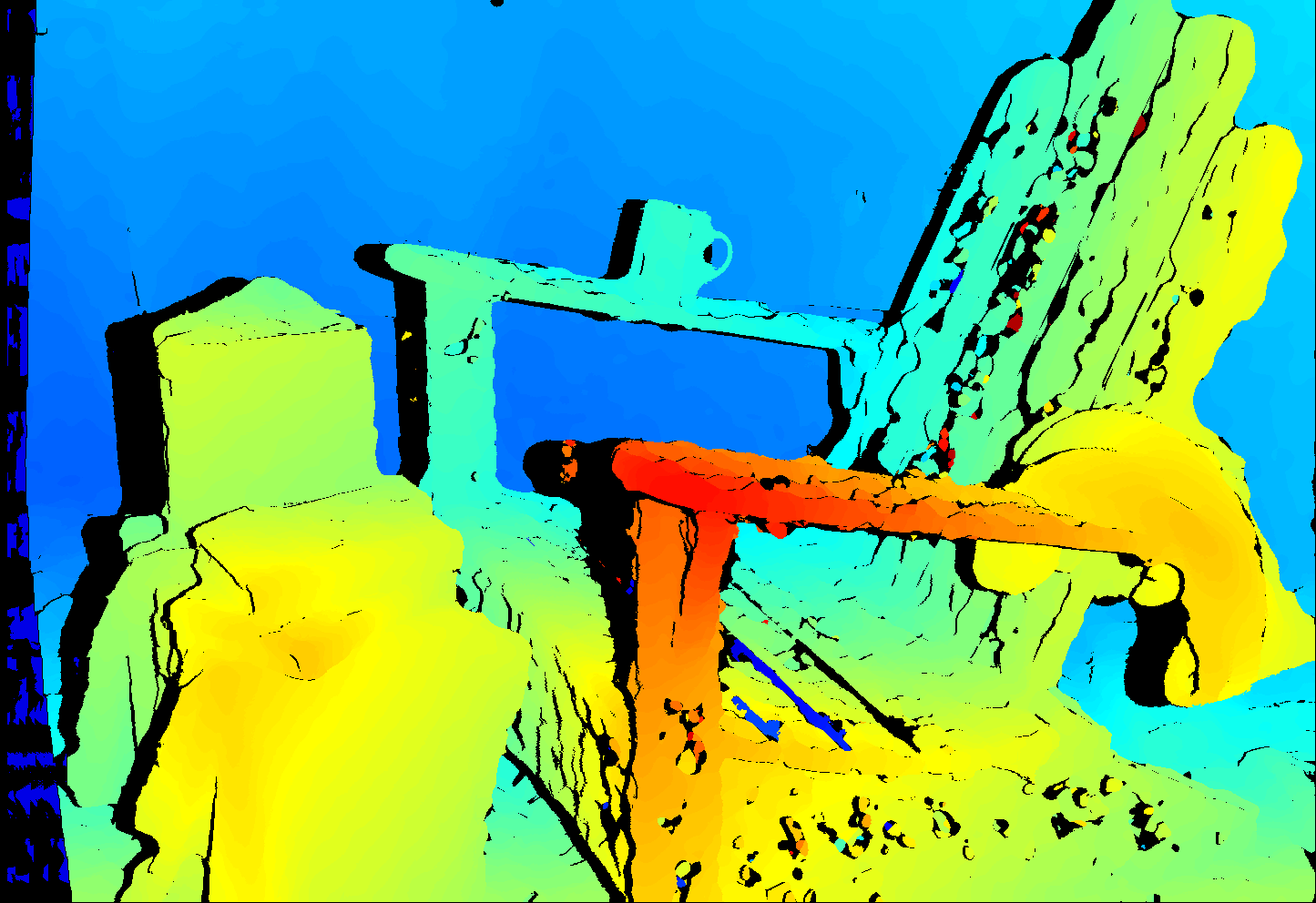}
   \includegraphics[width=0.45\linewidth]{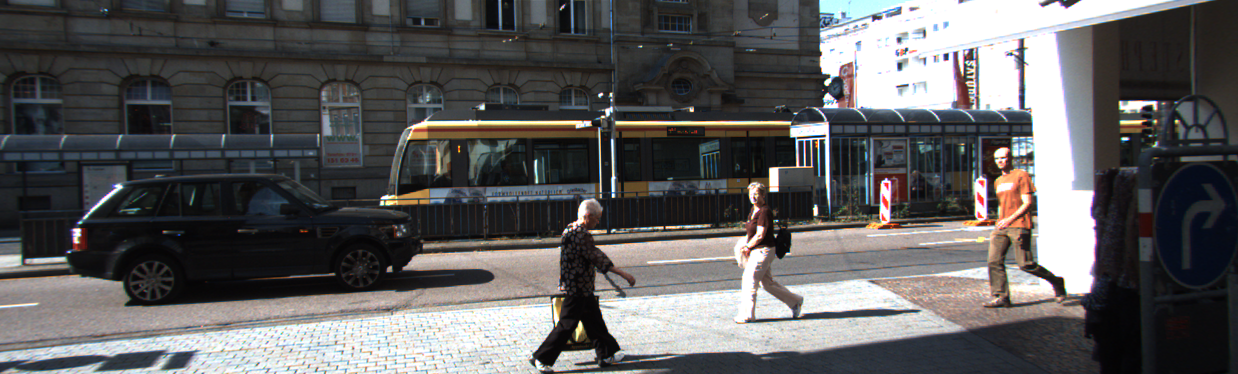}
    \includegraphics[width=0.45\linewidth]{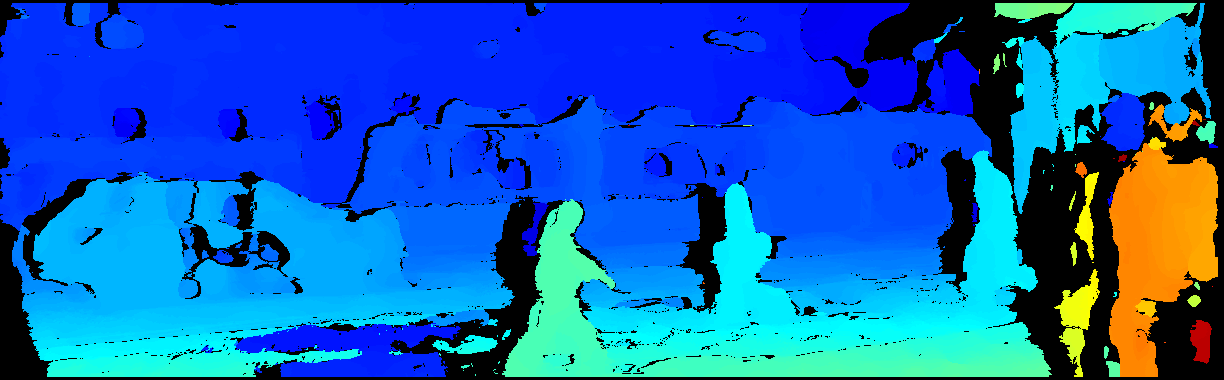}

   \caption{Qualitative results of our method. From left to right, top to bottom: left image of an indoor scene from the Middlebury dataset, the resulting disparity map after training, left image of a driving scene from the KITTI2015 dataset, the resulting disparity map after training.}
   \label{fig:disp_mb_kitti}

\end{figure}

Tab.~\ref{tab:epe_mb_kitti} shows different end-point errors of our method for Middlebury and KITTI2015.
Fig.~\ref{fig:disp_mb_kitti} shows the left image and the predicted disparity map of our method for this image for one example of the Middlebury dataset and one example of the KITTI2015 dataset after training. 

\section{Conclusion}

In this work, we have presented a fully self-supervised adaptive stereo method based on deep learning for remote-sensing application we call \textbf{S}elf-Supervised \textbf{ad}aptive convolutional neural \textbf{net}work or in short SAda-Net. We introduce a novel self-supervised training method which is based on adaptively updating the ground truth that is created by our network in each training step. To this end, we remove noisy and incorrect points from the map using the left-right consistency check.
We argued for the feasibility of tracking inconsistent points in order to track the training process and network convergence if no other information is present. We then evaluated our method on a challenging real-life satellite imagery from the WorldView-3 satellite. We have shown that our method is able to compete with other state-of the art methods. While fine-tuning the trained weights of the evaluated methods on the new scenes can improve the accuracy of the specific method, we argue that accurate ground truth is often missing and too expensive to create. We therefore argue that our method, that does not rely on costly ground truth data but rather can use any satellite imagery for training is a step towards truly autonomous stereo vision for remote sensing.

\end{document}